# An Efficient System for Automatic Map Storytelling – A Case Study on Historical Maps

Ziyi Liu*
ziyliu@student.ethz.ch
Institute of Cartography and
Geoinformation, ETH Zurich
Switzerland

Claudio Affolter*
claudaff@student.ethz.ch
Institute of Cartography and
Geoinformation, ETH Zurich
Switzerland

Sidi Wu†
sidiwu@ethz.ch
Institute of Cartography and
Geoinformation, ETH Zurich
Switzerland

Yizi Chen
yizi.chen@ethz.ch
Institute of Cartography and
Geoinformation, ETH Zurich
Switzerland

Lorenz Hurni
lhurni@ethz.ch
Institute of Cartography and
Geoinformation, ETH Zurich
Switzerland

## ABSTRACT

Historical maps provide valuable information and knowledge about the past. However, as they often feature non-standard projections, hand-drawn styles, and artistic elements, it is challenging for non-experts to identify and interpret them. While existing image captioning methods have achieved remarkable success on natural images, their performance on maps is suboptimal as maps are underrepresented in their pre-training process. Despite the recent advance of GPT-4 in text recognition and map captioning, it still has a limited understanding of maps, as its performance wanes when texts (e.g., titles and legends) in maps are missing or inaccurate. Besides, it is inefficient or even impractical to fine-tune the model with users' own datasets. To address these problems, we propose a novel and lightweight map-captioning counterpart. Specifically, we fine-tune the state-of-the-art vision-language model CLIP to generate captions relevant to historical maps and enrich the captions with GPT-3.5 to tell a brief story regarding *where*, *what*, *when* and *why* of a given map. We propose a novel decision tree architecture to only generate captions relevant to the specified map type. Our system shows invariance to text alterations in maps. The system can be easily adapted and extended to other map types and scaled to a larger map captioning system. The code is open-sourced at https://github.com/claudaff/automatic-map-storytelling.

## 1 INTRODUCTION

Historical maps allow us to learn more about a certain place's geography, economics, history, and culture. However, unlike modern maps, they often contain less accurate geographic information, varying artistic or religious symbols and legends, non-standard projections, and hand-drawn styles. This challenges non-experts (i.e., non-cartographers) to correctly identify and capture the key information. Image captioning [1, 2, 8, 9] provides descriptions for images in natural language and serves as a powerful tool in various situations, such as content understanding for individuals with visual impairments, image tagging for database management, and efficient search and retrieval of images. Typically, an image encoder is trained for visual cues, and a textual decoder is used to produce the final caption. CLIP (Contrastive Language-Image Pre-Training), recently proposed by [6], learns the shared representations for images and text prompts. It was trained over a tremendous number of images for a good correlation between images and texts and has been widely used for downstream tasks (like image captioning) with little or no further training. For example, the ClipCap model [5] uses the pre-trained CLIP prefix and fine-tunes a language model to generate image captions, which has achieved state-of-the-art performance. However, most image captioning methods generate descriptions limited to visual elements, which are not sufficient to tell a meaningful story about maps.

In this paper, we propose a map-specific captioning system equipped with a basic understanding of maps, which is not yet addressed by any image-captioning models. By fine-tuning CLIP models for map-relevant captions and using GPT (Generative Pre-trained Transformer) to combine and enrich them, our system could generate a comprehensive story. Given an input map, the story should answer the following questions: *where* does the map depict about? *what* is the map type, style, and topic? *when* was the map created? *why* was the map created? We choose GPT-3.5-turbo which can generate a more rapid response with equivalent performance for this task and a much lower cost, compared with GPT-4.

We focus on two major map types: topographic maps, which provide detailed and accurate graphical representations of an area [3], and pictorial maps, which use illustrations to represent information [7]. Since not every aspect is relevant for both map types — for example, the topic is usually the same for all the topographic maps (i.e., geography and topography) — we propose a novel decision tree structure to generate the captions with respect to the map type. Moreover, we design a user interface for interactive map storytelling, where the user can choose which aspects to include in the story, as shown in Figure 1.

## 2 METHODS AND IMPLEMENTATION

An overview of our methods is presented in Figure 1. We first process maps and their metadata automatically from the online map repository to generate a training dataset with keyword captions regarding *where*, *what* and *when* and use this dataset to fine-tune different CLIP models. In the inference phase, we propose a decision tree architecture to structure the keyword captions with respect to the map type and use GPT to extend the context (*why*) and summarize the story. Furthermore, a web interface is developed for interactive storytelling with the decision tree architecture and fine-tuned models loaded at the backend.

### 2.1 Dataset preparation

We collected data from the David Rumsey Historical Map Collection[1], an online map repository containing historical maps from all over the world complemented with detailed metadata. As we focus on topographic maps and pictorial maps, only the maps in the collection's categories *Classical* and *Pictorial map* were considered. In total, after manually filtering out poor-quality maps, 1,334 topographical and 3,183 pictorial maps were gathered.

To create ground-truth captions answering the four questions introduced in Section 1, we extracted necessary information from the metadata associated with each map. We processed topographic maps and pictorial maps separately as different challenges occurred.

*Where.* For topographic maps, since the location attribute in the metadata is often ambiguous, incorrect, and imprecise, we also parsed the location information from map titles. For pictorial maps, a substantial class imbalance emerged, with 3,183 maps depicting 1,349 different locations. Consequently, we decided to only focus on the two largest classes – the world and the United States.

*What.* For topographic maps, there are a few style variations, such as *with/without relief*, *with/without decorative elements*, and *hand-colored/engraved*, often described in metadata. However, as this description is not well structured and consistent, we have only extracted keywords from these descriptions. We calculated the frequencies of each keyword and then reduced the number of style classes to focus only on the most frequent ones. As topographic maps mainly describe the geography and topography of an area, we omitted the map topic in the caption. Pictorial maps are less constrained in styles with diverse color schemes and artistic illustrations, making it challenging to summarize the style. Thus, we excluded styles when captioning pictorial maps. Similar to *where*, the topics of pictorial maps present a strong imbalance. For example, there are 29% flight network maps but only 2% military maps. We decided to focus only on the most frequent topics and manually merged some sub-categories into a more general class.

*When.* We derived the century of production from the metadata. However, as most pictorial maps were created in the 20$^{th}$ century, it was no longer necessary to depict when they were created in the caption.

*Why.* The metadata provides no information about the purpose and functionality of a map. To fill this gap, we made use of GPT's generative capabilities. Instead of using the generated caption as ground truth to fine-tune the model, which would take additional training effort and might lead to error propagation stemming from imperfect captions, we only made use of GPT in the inference step.

We prepare separate datasets for different caption categories. Each dataset contains maps (compressed, up to 768×768 pixels) and the corresponding captions. Table 1 gives an overview of the final number of classes and maps for each category.

Table 1: Overview of the generated datasets for each caption category. Both numbers of classes and map samples are shown. We differentiate the *location* for topographic and pictorial maps.

| Caption category | Map type | Topographic | | | Pictorial | |
|---|---|---|---|---|---|---|
| | | Location (topo.) | Style | Century | Location (pict.) | Topic |
| # classes | 2 | 27 | 6 | 4 | 2 | 13 |
| # of maps | 4517 | 723 | 1132 | 1334 | 290 | 284 |

### 2.2 Fine-tuning CLIP

The visual information from maps is captured and transformed into textual information using CLIP models, each fine-tuned to generate the keyword caption for a specific aspect. We utilize six CLIP models in total, generating keyword captions related to location, map type, topic, style, and century, as shown in Figure 1 A). The fine-tuning process was adapted from [6]. We used a batch size of 10 and an initial learning rate of 1e-5 with Adam optimizer [4]. All models were trained on a single 16 GB NVIDIA RTX A4000 GPU.

### 2.3 Decision tree for inference

As some aspects are only relevant to certain map types, we proposed a decision tree structure where our models first predict the map type at the root node and then generate relevant keyword captions based on the identified type. For instance, given the map on the left in Table 3, the decision tree classifies it as a "pictorial map" (keyword 1), leading to the prediction of only the location "world" (keyword 2) and the topic "flight network" (keyword 3), while the style keyword is excluded as it is irrelevant in this context (see Section 2.1). At last, we use GPT to extend the story about *why* based on the generated keyword captions and to summarize the story by answering the questions in Section 1, with the prompt of the following structure: *"Please create a concise sentence that encapsulates these keywords:{keywords}. Additionally, provide a brief explanation in under 30 words, about {questions}"*.

## 3 RESULTS

### 3.1 Fine-tuned CLIP Models

We compare the prediction accuracy of our fine-tuned CLIP models with the base CLIP model. The base CLIP model can predict never-seen classes as long as the enumeration of class names is given. The similarity between the text encoding (class name) and the image encoding is then used to predict the most probable class. As shown in Table 2, based on 113 test maps (68 topographic maps and 45 pictorial maps). our fine-tuned CLIP models significantly outperform the base model in five out of six caption categories. The base model performed slightly better in the *location (pictorial)* caption category, likely due to its extensive training on illustrations of the United States and the world with significant graphic variations.

---
[1]https://www.davidrumsey.com/

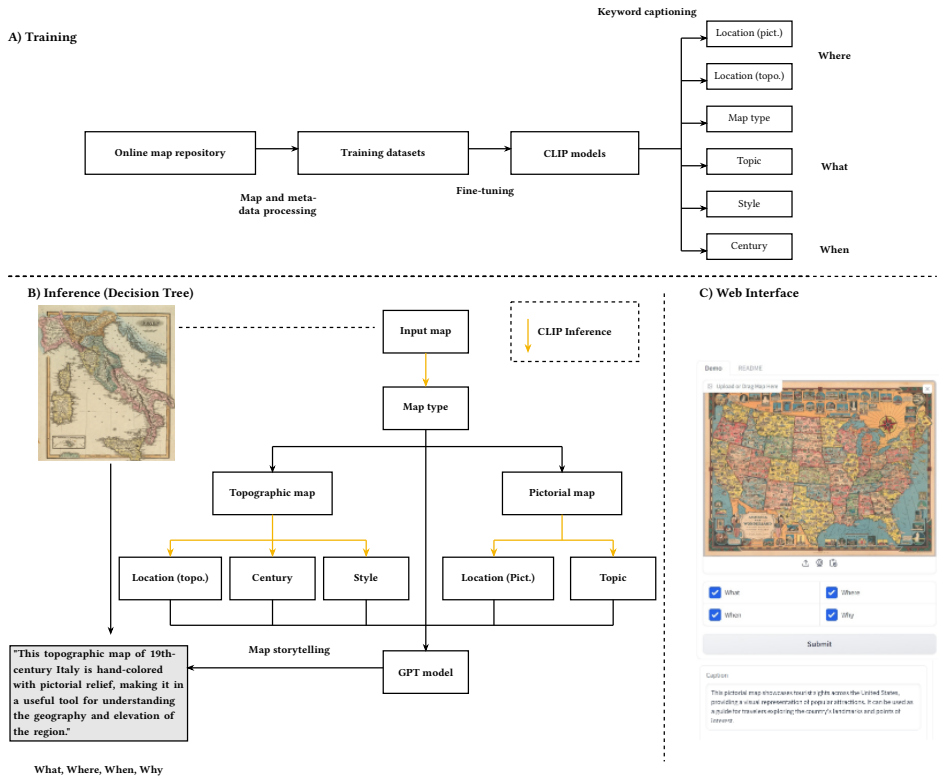

Figure 1: Overview of our proposed methodology.

Table 2: Comparison of prediction accuracies achieved per caption category between the base CLIP model and our fine-tuned CLIP models.

| Caption category | Map Type | Location (topo.) | Style | Century | Location (pict.) | Topic | Ave. Acc. |
|---|---|---|---|---|---|---|---|
| Base CLIP | 0.43 | 0.28 | 0.29 | 0.40 | **0.96** | 0.47 | 0.47 |
| Fine-tuned CLIP | **0.96** | **0.78** | **0.75** | **0.76** | 0.93 | **0.67** | **0.81** |

## 3.2 Map captioning

We compared our map captioning system with ClipCap and GPT-4. Table 3 shows examples of the stories generated by our method and ClipCap. While the original ClipCap can recognize maps, there are wrong interpretations like "room" and "map cutter". To fine-tune it, we combine *topic* and *location* for pictorial maps and *century* and *location* for topographic maps in a single sentence. In Table 3, while the fine-tuned ClipCap correctly detects the Air France global flight network in the first example, it falsely recognizes the production time (should be 19[th] instead of 17[th] century) in the second example. By comparison, our method can generate more accurate, comprehensive, and detailed captions, including *what*, *when*, *where*, and *why*. In Table 4, while GPT-4 presents superior capability in recognizing texts in maps, it does not understand the depicted geography (mainly France) and fails completely when the title texts are occluded, while our method shows rather stable performance.

## 4 CONCLUSION AND OUTLOOK

While existing image captioning methods show promising results on natural images, their performances for maps remain suboptimal in terms of caption accuracy and granularity. Our proposed method outperforms ClipCap in map storytelling and is more stable than GPT-4 when texts in maps are missing or altered. Compared with GPT-4, our proposed lightweight method can be easily used to fine-tune map captioning with users' private or proprietary datasets. Moreover, our system has a scalable decision tree architecture that is flexible to adapt and extend. However, there are also limitations. The current system focuses on broad periods (e.g., centuries) for identifying *when*, which can fail to capture significant historical nuances. Additionally, the caption quality depends on the current language model's capabilities, which may lack depth in explaining the *why* behind a map. In the future, more efficient ways can be explored to automatically generate a larger and more diverse map dataset. Moreover, the caption quality can be further strengthened. In combination with our decision tree approach, it would allow the development of a more powerful (historical) map captioning system.

## REFERENCES
[1] Peter Anderson, Xiaodong He, Chris Buehler, Damien Teney, Mark Johnson, Stephen Gould, and Lei Zhang. 2018. Bottom-up and top-down attention for image captioning and visual question answering. In *Proceedings of the IEEE conference on computer vision and pattern recognition*, 6077–6086.
[2] Xinlei Chen and C Lawrence Zitnick. 2014. Learning a recurrent visual representation for image caption generation. *arXiv preprint arXiv:1411.5654*.

Table 3: The stories of the same test maps generated by our method and ClipCap.✼: Fine-tuned ClipCap.

| | | |
|---|---|---|
| Test map | 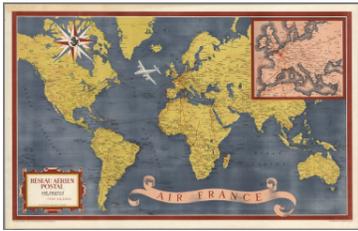 | 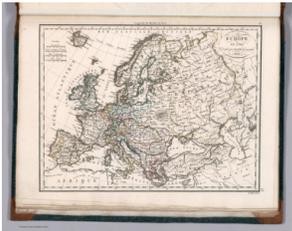 |
| Ours | This pictorial map illustrates the global flight network, showcasing worldwide destinations and travel routes. It is a visual representation of the world, providing information about flight connections, and can be used for planning and visualizing travel itineraries. | This hand-colored topographic map of Europe in the 19th century features pictorial relief. It shows the geographical features of Europe and can be used for geographical analysis. |
| ClipCap [5] | A map of the world is on display in a room. | An old map with a map cutter on it. |
| ClipCap✼ [5] | Map depicting Air France worldwide flight network. | Map depicting Europe in the 17th century. |

Table 4: Comparison of generated captions between our system and GPT-4 influenced by texts. On the right, the title and production year (at bottom-left) are covered.

| | | |
|---|---|---|
| Test map | 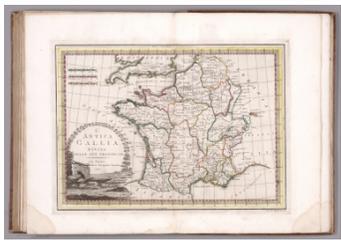 | 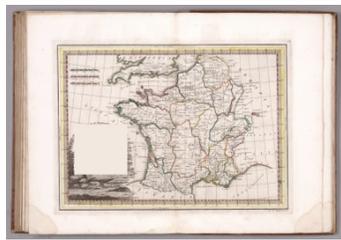 |
| Ours | A topographic map of 18th century France with decorative elements and pictorial relief, useful for visualizing the landscape features and terrain of the region. | A decorative topographic map of 18th century France with pictorial relief elements. This map showcases the geographical features of France through artistic design and can be used for historical reference and decoration. |
| GPT-4 | This map depicts "Antica Gallia," or ancient Gaul, detailing its divisions during Roman times. It was likely created for historical and educational purposes in the 19th century. | This is a historical map of the Indian subcontinent, likely created in the 19th century, used for educational and colonial administrative purposes. |

[3] Alexander Kent. 2009. Topographic maps: methodological approaches for analyzing cartographic style. *Journal of Map & Geography Libraries*, 5, 2, 131–156.
[4] Diederik P Kingma and Jimmy Ba. 2014. Adam: a method for stochastic optimization. *arXiv preprint arXiv:1412.6980*.
[5] Ron Mokady, Amir Hertz, and Amit H Bermano. 2021. Clipcap: clip prefix for image captioning. *arXiv preprint arXiv:2111.09734*.
[6] Alec Radford et al. 2021. Learning transferable visual models from natural language supervision. PMLR, (2021).
[7] Raimund Schnürer, Rene Sieber, Jost Schmid-Lanter, A Cengiz Öztireli, and Lorenz Hurni. 2021. Detection of pictorial map objects with convolutional neural networks. *The Cartographic Journal*, 58, 1, 50–68.
[8] Matteo Stefanini, Marcella Cornia, Lorenzo Baraldi, Silvia Cascianelli, Giuseppe Fiameni, and Rita Cucchiara. 2022. From show to tell: a survey on deep learning-based image captioning. *IEEE transactions on pattern analysis and machine intelligence*, 45, 1, 539–559.
[9] Luowei Zhou, Hamid Palangi, Lei Zhang, Houdong Hu, Jason Corso, and Jianfeng Gao. 2020. Unified vision-language pre-training for image captioning and vqa. In *Proceedings of the AAAI conference on artificial intelligence*. Vol. 34, 13041–13049.